\documentclass[letterpaper, 10 pt, conference]{ieeeconf}  
\IEEEoverridecommandlockouts             
\makeatletter
\let\NAT@parse\undefined
\makeatother
\usepackage[export]{adjustbox}
\usepackage{hyperref}
\usepackage{xurl}
\usepackage[linesnumbered,ruled,vlined,algo2e]{algorithm2e}
\usepackage{amsfonts}
\usepackage{amsmath}
\usepackage{amssymb}
\usepackage{xcolor}
\usepackage{mathtools}
\usepackage{graphicx}
\usepackage{subcaption}
\usepackage{multirow}
\usepackage{cite}
\usepackage{mathrsfs}
\usepackage{acronym}
\usepackage{bm}
\usepackage{stackengine}
\usepackage{siunitx}

\usepackage{amsthm}
\usepackage[T1]{fontenc}

\theoremstyle{plain}

\theoremstyle{definition}

\newtheorem{problem}{Problem}

\theoremstyle{remark}

\title{ \LARGE \bf
  Onboard Vision and MPC Navigation for Underwater Robots: An Open BlueROV2 Platform for Multi-Robot Experiments \& Docking
}

\author{
Victor Nan Fernandez-Ayala$^1$, Wiktor Kowalczyk$^{1,*}$, Cezary Banaszek$^{1,*}$ and Dimos V. Dimarogonas$^1$
\thanks{$^*$Equal contribution.}
\thanks{$^1$Department of Decision and Control Systems, School of Electrical Engineering and Computer Science, KTH Royal Institute of Technology, Stockholm, Sweden. 
Emails: \texttt{\{vnfa, wmko, banaszek, dimos\}@kth.se}}
\thanks{
This work was supported by Vinnova, Sweden's Innovation Agency, within the Advanced Digitalisation programme (grant no. 2024-01430) with co-funding from KTH Digital Futures and the Swedish Defence Materiel Administration (FMV).}
}

\begin{document}
\maketitle


\begin{abstract}
Autonomous underwater robots require robust perception, estimation and control to operate in confined environments. This paper presents an open-source BlueROV2 platform combining onboard vision with nonlinear Model Predictive Control (NMPC) for autonomous navigation and docking. The platform integrates an NVIDIA Jetson Orin NX and an Intel RealSense D435i stereo camera in a modular pressure housing. Underwater-calibrated stereo depth and real-time object detection provide relative position measurements of nearby BlueROV2 vehicles in the camera and body frames. A quaternion-based estimator fuses external pose and inertial measurements, while an NMPC controller based on a nonlinear six-degree-of-freedom model tracks planned navigation and docking trajectories. To support reproducible development, we also provide open-source physics-based PX4 SITL and Gazebo environments, multi-robot simulation tools and a low-cost physical docking station. Experiments evaluate underwater perception, onboard computational performance, state estimation, trajectory tracking and autonomous docking.
\end{abstract}


\section{Introduction} \label{sec:introduction}

Autonomous underwater robots can reduce the cost and risk of marine inspection, environmental monitoring, scientific exploration and infrastructure maintenance. However, their deployment remains challenging due to the coupled nonlinear vehicle dynamics, limited communication, uncertain localization and optical degradation from light attenuation, scattering and refraction \cite{yuh2000design,jordt2012refractive}. These difficulties are particularly relevant in confined environments, where accurate perception, state estimation and control operate concurrently with limited onboard computational resources.

Low-cost open-source underwater platforms have considerably improved access to experimental marine robotics. The LoCO AUV provides an open and modular vision-guided platform \cite{edge2020loco}, while BlueROV2-based systems showed how a commercially available Remotely Operated Vehicle (ROV) can be extended for autonomous operation \cite{wilby2020lowcost}. Open simulation models have also been developed to support control design for the BlueROV2 \cite{vonbenzon2022benchmark}. Nevertheless, the implementation of a complete autonomy stack still requires the integration of embedded computing, perception, state estimation, planning, control and suitable experimental infrastructure.

Vision is especially useful for short-range underwater navigation, since cameras provide geometric and semantic information using compact and comparatively inexpensive sensors. Underwater camera measurements, however, require a suitable calibration to account for the refractive interfaces introduced by the camera housing and the outside water \cite{jordt2012refractive}. During docking, visual markers can provide accurate relative pose measurements for the final approach and have been previously combined with other localization sources in autonomous underwater docking systems \cite{creutz2023docking}. Nonlinear Model Predictive Control (NMPC) is also well suited for this type of operation, since it accounts for nonlinear vehicle dynamics and actuator constraints while re-optimizing from the latest state estimate at every  sampling instant, allowing feedback correction of disturbances and model mismatch \cite{fernandez2017mpc}.

\begin{figure}[t]
    \centering
    \includegraphics[width=\columnwidth]{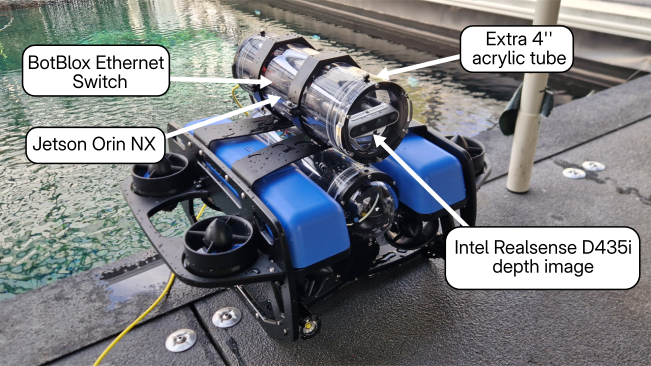}
    \caption{BlueROV2 Heavy with the modular autonomy tube.}
    \label{fig:autonomous_bluerov2}
\end{figure}

In this work, we develop an open autonomy platform for the BlueROV2 Heavy that combines embedded stereo perception, quaternion-based state estimation, reference planning, NMPC navigation and docking. All perception, estimation, planning and control computations are executed onboard. 
The platform builds upon the BlueROV2 fleet and six-degree-of-freedom (DoF) model of the Marinarium, an underwater robotics test facility introduced in \cite{torroba2026marinariumnewarenabring}, where it is experimentally evaluated using underwater motion capture. The resulting platform is shown in Fig.~\ref{fig:autonomous_bluerov2}.

The contributions are: (i) an onboard autonomy framework combining calibrated stereo relative localization, quaternion-based state estimation, reference planning and NMPC for navigation and docking; (ii) an open-source software and hardware platform implementing this stack on the BlueROV2, together with a low-cost docking station, PX4 SITL/Gazebo environments and multi-robot simulation tools; and (iii) experimental evaluation of underwater stereo, object detection, onboard computation, state estimation, NMPC tracking and autonomous docking.


\section{Vehicle Model and Problem Formulation}
\label{sec:preliminaries}

\subsection{Vehicle Model}

Consider the modified BlueROV2 Heavy shown in Fig.~\ref{fig:autonomous_bluerov2}. The vehicle has an additional tube with the onboard autonomy hardware, which changes its total mass, displaced volume and inertia. We denote the North-East-Down (NED) navigation frame by $\{n\}$ and the Forward-Right-Down (FRD) body-fixed frame by $\{b\}$. In the standard 6 DoF marine vehicle formulation \cite{fossen2021handbook}, the vehicle state is
\begin{equation}
    \bm{x}
    =
    \begin{bmatrix}
        (\bm{p}^{n})^\top &
        \bm{q}^\top &
        \bm{\nu}^\top
    \end{bmatrix}^{\top}
    \in \mathbb{R}^{3}\times\mathbb{S}^{3}\times\mathbb{R}^{6},
    \label{eq:state_definition}
\end{equation}
where $\bm{p}^{n}=[p_n,p_e,p_d]^\top$ is the position in $\{n\}$, $\bm{q}=[q_w,q_x,q_y,q_z]^\top$ is the unit quaternion representing the rotation from $\{b\}$ to $\{n\}$, and $\bm{\nu}=[\bm{v}^\top,\bm{\omega}^\top]^\top=[u,v,w,p,q,r]^\top$ contains the body-frame linear and angular velocities. The control input is the desired body wrench
\begin{equation}
    \bm{\tau}
    =
    \begin{bmatrix}
        \bm{f}^{\top} & \bm{m}^{\top}
    \end{bmatrix}^{\top}
    =
    [F_x,F_y,F_z,M_x,M_y,M_z]^\top \in \mathbb{R}^{6},
    \label{eq:wrench_definition}
\end{equation}
which is mapped to the eight individual thruster commands by the PX4 control allocation layer.

Let $\bm{R}(\bm{q})\in SO(3)$ denote the rotation matrix from $\{b\}$ to $\{n\}$ and let $\bm{q}_v=[q_x,q_y,q_z]^\top$. The vehicle kinematics are
\begin{align}
    \dot{\bm{p}}^{n}
        &= \bm{R}(\bm{q})\bm{v},
        \label{eq:position_kinematics}\\
    \dot{\bm{q}}
        &= \frac{1}{2}
        \begin{bmatrix}
            -\bm{q}_v^\top\\
            q_w\bm{I}_3+\bm{S}(\bm{q}_v)
        \end{bmatrix}
        \bm{\omega},
        \label{eq:quaternion_kinematics}
\end{align}
where $\bm{S}(\cdot)$ is the skew-symmetric matrix operator and $\bm{q}^{\top}\bm{q}=1$. The dynamics are then adapted from the BlueROV2 model in \cite{vonbenzon2022benchmark} and \cite{torroba2026marinariumnewarenabring}, but its parameters are modified to account for the mass and buoyancy of the additional autonomy tube. The dynamics are thus given by
\begin{equation}
    \dot{\bm{\nu}}
    =
    \bm{M}^{-1}\!\left(
        \bm{\tau}
        -\bm{C}(\bm{\nu})\bm{\nu}
        -\bm{D}(\bm{\nu}_r)\bm{\nu}_r
        -\bm{g}(\bm{q})
    \right),
    \label{eq:fossen_model}
\end{equation}
where $\bm{M}=\bm{M}_{RB}+\bm{M}_{A}$ are the rigid-body and added-mass contributions, respectively, and $\bm{C}(\bm{\nu})=\bm{C}_{RB}(\bm{\nu})+\bm{C}_{A}(\bm{\nu})$ contains the corresponding Coriolis and centripetal terms. A constant irrotational water current $\bm{v}^{n}_{c}$ can be included through the vehicle velocity relative to the surrounding water $\bm{\nu}_r=[(\bm{v}-\bm{R}^{\top}(\bm{q})\bm{v}^{n}_{c})^\top,\bm{\omega}^\top]^\top$. The damping matrix, with linear and quadratic components \cite{fossen2021handbook}, is written as
\begin{equation}
    \bm{D}(\bm{\nu}_r)
    =
    \operatorname{diag}
    \left(
        \bm{d}_{l}
        +
        \bm{d}_{q}\odot|\bm{\nu}_r|
    \right),
    \label{eq:damping_matrix}
\end{equation}
where $\odot$ denotes the Hadamard (element-wise) product. Let $m$, $V$, $\rho$ and $g$ denote the vehicle mass, volume, water density and gravitational acceleration, respectively. Then, $W=mg$, $B=\rho gV$ and $\bm{R}(\bm{q})=[r_{ij}]$. With the center of gravity located at the body-frame origin and the center of buoyancy $\bm{r}_b=[x_b,y_b,z_b]^\top$, the restoring vector used in the model is
\begin{equation}
\bm{g}(\bm{q})=
\begin{bmatrix}
 -(W-B)r_{31}\\
 -(W-B)r_{32}\\
 -(W-B)r_{33}\\
 B(y_b r_{33}-z_b r_{32})\\
 B(z_b r_{31}-x_b r_{33})\\
 B(x_b r_{32}-y_b r_{31})
\end{bmatrix}.
\label{eq:restoring_vector}
\end{equation}

The parameters for the vehicle with the autonomy tube are summarized in Table~\ref{tab:heavy_tube_parameters}. The mass and volume are obtained from the physical vehicle, with the inertia obtained by scaling the benchmark values in \cite{vonbenzon2022benchmark} by the vehicle mass ratio. These parameters result in $B-W=\SI{0.491}{N}$, yielding the intended near-neutral, slightly positively buoyant configuration.

\begin{table}[t]
    \centering
    \caption{Parameters of the modified BlueROV2 Heavy.}
    \label{tab:heavy_tube_parameters}
    \footnotesize
    \begin{tabular}{c|l}
        \hline
        Parameter & Value in SI units\\
        \hline
        $m$, $V$
        &
        $\SI{15.6}{kg}$,
        $\SI{0.01565}{m^3}$
        \\
        $\rho$, $g$
        &
        $\SI{1000}{kg/m^3}$,
        $\SI{9.82}{m/s^2}$
        \\
        $\bm{r}_b$
        &
        $[0,0,-0.01]^\top\,\si{m}$
        \\
        $[I_x,I_y,I_z]$
        &
        $[0.3000,0.2652,0.4272]\,\si{kg.m^2}$
        \\
        $\operatorname{diag}(\bm{M}_A)$
        &
        $[6.36,7.12,18.68,0.189,0.135,0.222]$
        \\
        $\bm{d}_l$
        &
        $[13.7,0,33.0,0,0.8,0]$
        \\
        $\bm{d}_q$
        &
        $[141,217,190,1.19,0.47,1.50]$
        \\
        \hline
    \end{tabular}
\end{table}

Combining \eqref{eq:position_kinematics}, \eqref{eq:quaternion_kinematics} and \eqref{eq:fossen_model}, the state $\bm{x}$ in \eqref{eq:state_definition} evolves according to the continuous-time system
\begin{equation}
    \dot{\bm{x}}=\bm{f}(\bm{x},\bm{\tau}),
    \label{eq:continuous_dynamics}
\end{equation}
where $\bm{f}$ denotes the nonlinear vector field defined by the preceding kinematics and dynamics. For prediction and simulation, \eqref{eq:continuous_dynamics} is discretized using a fixed-step fourth-order explicit Runge-Kutta method \cite{acados}
\begin{equation}
    \bm{x}_{k+1}
    =
    \mathcal{N}_{q}\!\left(
        \operatorname{RK4}(\bm{x}_k,\bm{\tau}_k,T_s)
    \right)
    =:\bm{f}_d(\bm{x}_k,\bm{\tau}_k),
    \label{eq:discrete_dynamics}
\end{equation}
where $\bm{f}_d$ denotes the discrete-time dynamics and
$\mathcal{N}_{q}$ replaces $\bm{q}$ by
$\bm{q}/\|\bm{q}\|_2$, leaving the remaining state unchanged.

\subsection{Problem Formulation}

At time step $k$, the vehicle receives noisy IMU, stereo-camera and external-pose measurements and a desired terminal pose $(\bm{p}^n_d,\bm{q}_d)$, representing either a navigation goal or a docking configuration. Let
\begin{equation}
    \mathcal{X}^{r}_k
    =
    \left\{
        \bm{x}^{r}_{k+\ell|k}
    \right\}_{\ell=0}^{N}
    \label{eq:reference_sequence}
\end{equation}
denote a reference sequence terminating at the desired pose, where $N$ is the prediction-horizon length.

\begin{problem}
Using these measurements, estimate the current state $\hat{\bm{x}}_k$, generate $\mathcal{X}^{r}_k$ from $\hat{\bm{x}}_k$ to the terminal pose and determine the body-wrench sequence $\{\bm{\tau}_{\ell|k}\}_{\ell=0}^{N-1}$ such that $\bm{x}_{0|k}=\hat{\bm{x}}_k$, \eqref{eq:discrete_dynamics} is satisfied for $\hat{\bm{x}}_k$, the predicted state tracks $\mathcal{X}^{r}_k$, and $\bm{\tau}_{\ell|k}\in\mathcal{U}$ for $\ell=0,\ldots,N-1$, where
$\mathcal{U}=[\underline{\bm{\tau}},\overline{\bm{\tau}}]$ is the nonempty actuator-limited wrench set. All computations must be performed onboard within $T_s$.
\end{problem}


\section{Open Platform, Perception and Simulation}
\label{sec:system}

\subsection{Autonomy Hardware and Software}

The standard BlueROV2 Heavy is extended with a modular pressure housing mounted on top of the vehicle, as shown in Fig.~\ref{fig:autonomous_bluerov2}. The housing contains an NVIDIA Jetson Orin NX 8GB, an Intel RealSense D435i stereo camera and a compact Gigabit Ethernet switch. The D435i is installed behind the transparent front end cap, while the Jetson and networking hardware are mounted on removable internal trays. The complete module is attached to the standard BlueROV2 roof racks using detachable enclosure clamps.

On the Jetson, ROS~2 nodes acquire and align the D435i measurements, perform YOLO/depth-based relative localization, run the EKF and reference planner, and solve the NMPC. The Ethernet switch connects the Jetson, the vehicle autopilot and the external network interface. The autopilot runs PX4 \cite{meier2015px4} and exchanges vehicle measurements and body-wrench setpoints with the onboard ROS~2 nodes. The external connection is retained for monitoring, data recording and delivery of the motion-capture measurements, but no external computer is needed to execute the autonomy stack.

The CAD files, Bill of Materials (BoM), assembly instructions and software installation procedure are provided in the open-source project repository\footnote{\url{https://github.com/KTH-DHSG/BlueROV-Setup}}. The modular design allows the autonomy system to be installed on an existing BlueROV2 without modifying any of its components.

\subsection{Underwater Stereo Perception}

The D435i provides synchronized color and stereo depth measurements. Since the camera is placed behind an acrylic interface and operated underwater, its stereo parameters are recalibrated after installation. The resulting depth image is aligned with the color image so that the depth and object-detection outputs share the same pixel coordinates.

A lightweight YOLO detector \cite{yolo} is fine-tuned to identify other BlueROV2 vehicles. Let $\Omega_{j,k}$ denote the image region associated with the detection of vehicle $j$ at time step $k$, and let $D_k(a,b)$ denote the aligned depth measurement at pixel $(a,b)$. Invalid measurements and depth outliers are removed from this region, resulting in the set of accepted samples
\begin{equation}
    \mathcal{Z}_{j,k}
    =
    \left\{
        D_k(a,b):
        (a,b)\in\Omega_{j,k},
        \ D_k(a,b)\text{ accepted}
    \right\}.
    \label{eq:accepted_depth_set}
\end{equation}
The estimated depth of the detected vehicle is obtained as
\begin{equation}
    \hat{z}_{j,k}
    =
    \frac{1}{|\mathcal{Z}_{j,k}|}
    \sum_{z\in\mathcal{Z}_{j,k}}z.
    \label{eq:detected_object_depth}
\end{equation}

\begin{figure}[t]
    \centering
    \includegraphics[width=0.88\columnwidth]{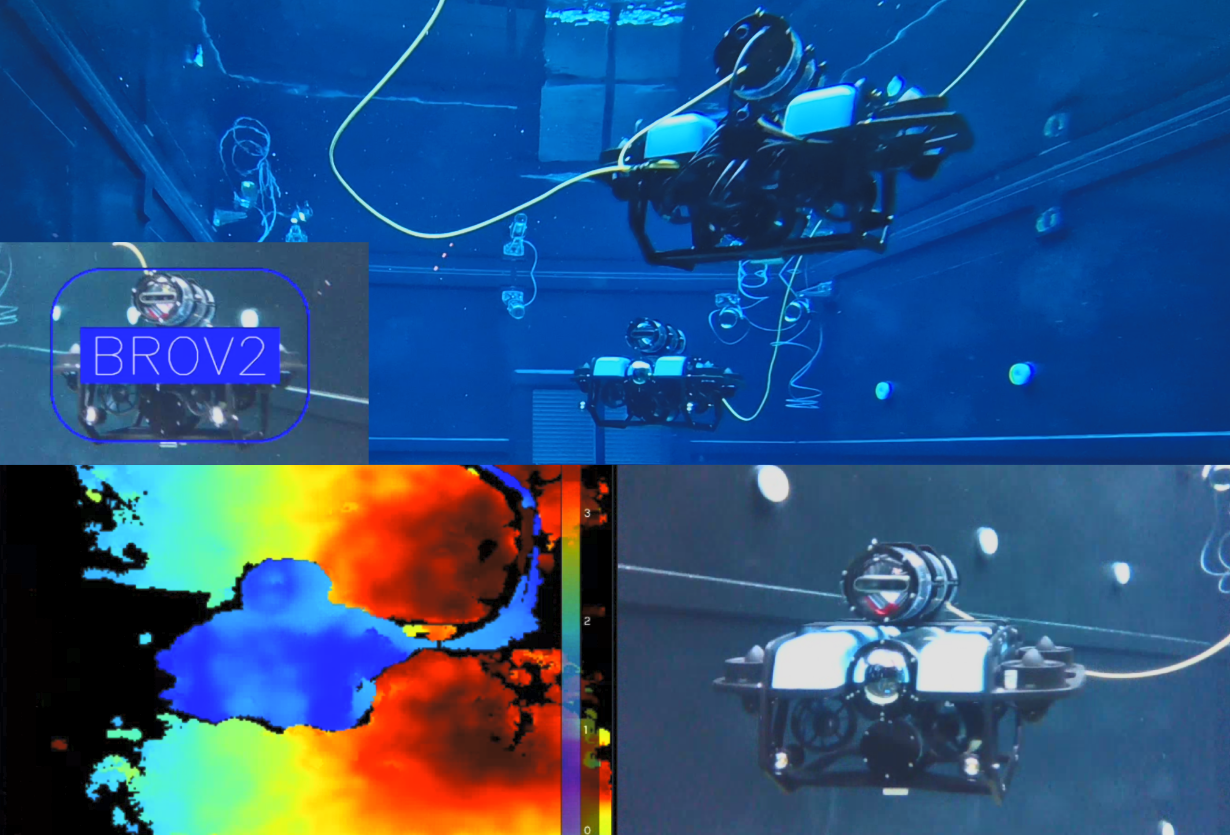}
    \caption{Underwater BlueROV2 detection and aligned stereo-depth output produced by the onboard perception module.}
    \label{fig:depth_yolo}
\end{figure}

Let $(\bar{a}_{j,k},\bar{b}_{j,k})$ be the center of the detected image region and let $\bm{K}$ be the calibrated intrinsic matrix \cite{jordt2012refractive}. The relative position of detected vehicle $j$ with respect to observer vehicle $i$ in the camera frame $\{c\}$ is calculated through back-projection as
\begin{equation}
    \hat{\bm{p}}^{c}_{j/i,k}
    =
    \hat{z}_{j,k}\bm{K}^{-1}
    \begin{bmatrix}
        \bar{a}_{j,k} &
        \bar{b}_{j,k} &
        1
    \end{bmatrix}^{\top}.
    \label{eq:camera_relative_position}
\end{equation}
Using the calibrated camera-to-body rotation ${}^{b}\bm{R}_{c}$ and translation ${}^{b}\bm{t}_{c}$, the corresponding body-frame measurement is
\begin{equation}
    \hat{\bm{p}}^{b}_{j/i,k}={}^{b}\bm{R}_{c}\hat{\bm{p}}^{c}_{j/i,k}
    +
    {}^{b}\bm{t}_{c}.
    \label{eq:body_relative_position}
\end{equation}
Thus, the relative position comes only from onboard measurements and does not need communication or global position of the vehicle. The trained YOLO model, ROS~2 modules for synchronized color and depth processing, and relative position estimation are released in the open-source perception repository\footnote{\url{https://github.com/KTH-DHSG/BlueROV-ROS-Modules}}. Figure~\ref{fig:depth_yolo} shows a snapshot of the video with the object-detection and aligned-depth outputs (see accompanying \href{https://youtu.be/VsCCucKPt8k}{video}). The perception module also supports ArUco-based relative pose estimation for docking \cite{aruco}.

\subsection{Simulation and Docking Infrastructure}

A physics-based simulation environment is developed using PX4 SITL and Gazebo \cite{meier2015px4}. The simulated BlueROV2 contains the complete eight-thruster configuration and reproduces the mass, buoyancy, inertia and hydrodynamic parameters of the modified vehicle. The simulation exposes the same PX4 and ROS~2 interfaces used by the physical platform, allowing the same estimation, planning and control nodes to operate in simulation and in the tank.

\begin{figure*}[t]
    \centering
    \begin{subfigure}[t]{0.275\textwidth}
        \centering
        \includegraphics[width=0.95\linewidth]{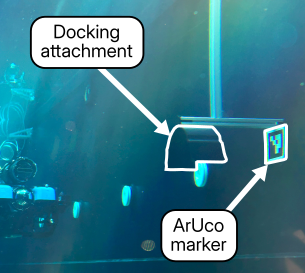}
        \caption{Docking station in the Marinarium.}
        \label{fig:physical_docking_station}
    \end{subfigure}
    \hfill
    \begin{subfigure}[t]{0.718\textwidth}
        \centering
        \includegraphics[width=\linewidth]{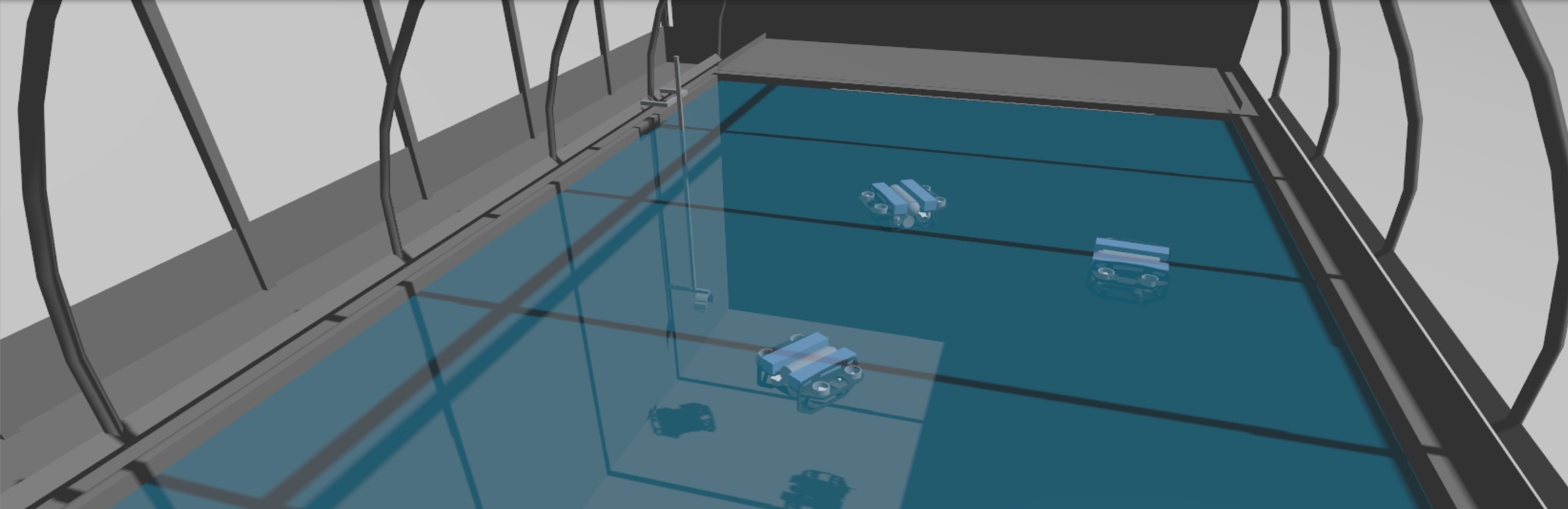}
        \caption{Multi-robot PX4 SITL simulation in the Marinarium environment (see accompanying \href{https://youtu.be/VsCCucKPt8k}{video}).}
        \label{fig:multi_robot_simulation}
    \end{subfigure}
    \caption{Open experimental infrastructure. The physical docking station and the simulator from the multi-robot launch file.}
    \label{fig:docking_and_simulation}
\end{figure*}

An underwater environment is also provided imitating the geometry of the Marinarium \cite{torroba2026marinariumnewarenabring}, including the tank boundaries and a simulated version of a docking station, as shown together with the physical one in Fig.~\ref{fig:docking_and_simulation}. An additional multi-vehicle launch file can be used to create several independent PX4 instances, with each robot assigned its own system identifier, ROS~2 namespace and communication topics.

The physical docking station is constructed from standard aluminium profiles and 3D-printed PLA components coated with a capillary sealant. It is detachable from the tank structure and does not require custom-machined components. It can also be equipped with a 3D printed ArUco marker. The same geometry and marker placement are reproduced in Gazebo. Its CAD files are also provided in the setup repository introduced above. The PX4 software developments and the Gazebo models and environments have already been merged upstream into the official PX4. Installation and execution instructions are provided in the setup repository.


\section{State Estimation and NMPC Navigation}
\label{sec:navigation}

\subsection{Quaternion-Based State Estimation}

The underwater motion-capture measurements contain noise and occasional outliers due to refraction and marker occlusion. A quaternion error-state Extended Kalman Filter (EKF) is thus used to fuse the measured global pose with the onboard accelerometer and gyroscope data \cite{markley2003attitude,crassidis2007survey}.

Let $\tilde{\bm{a}}^b_k$ and $\tilde{\bm{\omega}}^b_k$ denote the specific-force and angular-velocity measurements from the onboard Inertial Measurement Unit (IMU). They satisfy
\begin{align}
    \tilde{\bm{a}}^b_k
    &=
    \bm{R}^{\top}(\bm{q}_k)
    \left(
        \dot{\bm{v}}^n_k-\bm{g}^n
    \right)
    +\bm{b}_{a,k}+\bm{n}_{a,k},
    \label{eq:imu_acceleration_model}\\
    \tilde{\bm{\omega}}^b_k
    &=
    \bm{\omega}^b_k+\bm{b}_{g,k}+\bm{n}_{g,k},
    \label{eq:imu_gyro_model}
\end{align}
where $\bm{g}^n=[0,0,g]^\top$ in NED frame, $\bm{b}_{a,k}$ and $\bm{b}_{g,k}$ are the accelerometer and gyroscope biases, and $\bm{n}_{a,k}$ and $\bm{n}_{g,k}$ are zero-mean measurement noise terms. The estimator state is then $\bm{x}^e_k=[(\bm{p}^n_k)^\top,(\bm{v}^n_k)^\top, \bm{q}_k^\top,\bm{b}_{a,k}^\top,\bm{b}_{g,k}^\top]^\top$.


Let $\Delta t_k$ denote the IMU sampling time interval. Between global-pose updates, the IMU measurements propagate the state estimate at the IMU rate according to
\begin{align}
    \hat{\bm{a}}^n_k
    &=
    \bm{R}(\hat{\bm{q}}_k)
    \left(
        \tilde{\bm{a}}^b_k-\hat{\bm{b}}_{a,k}
    \right)
    +\bm{g}^n,
    \nonumber\\
    \hat{\bm{p}}^{n-}_{k+1}
    &=
    \hat{\bm{p}}^n_k
    +\Delta t_k\hat{\bm{v}}^n_k
    +\frac{\Delta t_k^2}{2}\hat{\bm{a}}^n_k,
    \nonumber\\
    \hat{\bm{v}}^{n-}_{k+1}
    &=
    \hat{\bm{v}}^n_k
    +\Delta t_k\hat{\bm{a}}^n_k,
    \nonumber\\
    \hat{\bm{q}}^-_{k+1}
    &=
    \mathcal{N}_q\left(
        \hat{\bm{q}}_k
        \otimes
        \operatorname{Exp}_q
        \left(
            \Delta t_k
            \left(
                \tilde{\bm{\omega}}^b_k
                -\hat{\bm{b}}_{g,k}
            \right)
        \right)
    \right),
    \label{eq:ekf_nominal_prediction}\\
    \hat{\bm{b}}^-_{a,k+1}
    &=
    \hat{\bm{b}}_{a,k},
    \qquad
    \hat{\bm{b}}^-_{g,k+1}
    =
    \hat{\bm{b}}_{g,k},
    \nonumber
\end{align}
where $\otimes$ denotes quaternion multiplication and
$\operatorname{Exp}_q:\mathbb{R}^3\rightarrow\mathbb{S}^3$ denotes the quaternion exponential map \cite{crassidis2007survey}. The biases are modelled as random walks in the covariance propagation.


The associated 15-dimensional error state is $\delta\bm{x}^e_k=
[\delta\bm{p}_k^\top,\delta\bm{v}_k^\top,\delta\bm{\theta}_k^\top,
\delta\bm{b}_{a,k}^\top,\delta\bm{b}_{g,k}^\top]^\top$, where $\delta\bm{\theta}_k\in\mathbb{R}^3$ is the local attitude error. Its covariance is propagated as
\begin{equation}
    \bm{P}^{-}_{k+1}
    =
    \bm{F}_k\bm{P}_k\bm{F}_k^\top
    + \bm{G}_k\bm{Q}^e_k\bm{G}_k^\top,
    \label{eq:ekf_covariance_prediction}
\end{equation}
where $\bm{F}_k$ and $\bm{G}_k$ are the discrete error-dynamics Jacobians computed over $\Delta t_k$, and $\bm{Q}^e_k$ contains the IMU and bias random-walk covariances. Let $\bm{p}^m_k$ and $\bm{q}^m_k$ be a global pose measurement. The position and multiplicative attitude residuals are then
\begin{equation}
    \bm{r}_k
    =
    \begin{bmatrix}
        \bm{p}^m_k-\hat{\bm{p}}^{n-}_k\\
        \operatorname{Log}_q
        \left(
            \bm{q}^m_k
            \otimes
            (\hat{\bm{q}}^-_k)^{-1}
        \right)
    \end{bmatrix},
    \label{eq:ekf_measurement_residual}
\end{equation}
where $\operatorname{Log}_q:\mathbb{S}^3\rightarrow\mathbb{R}^3$ is the quaternion logarithmic map. Using error-state correction \cite{markley2003attitude,crassidis2007survey}, the EKF update is
\begin{align}
    \bm{K}_k
    &=
    \bm{P}^-_k\bm{H}_k^\top
    \left(
        \bm{H}_k\bm{P}^-_k\bm{H}_k^\top
        +\bm{\Sigma}_m
    \right)^{-1},
    \nonumber\\
    \delta\hat{\bm{x}}^e_k
    &=
    \bm{K}_k\bm{r}_k,
    \label{eq:ekf_state_correction}\\
    \bm{P}_k
    &=
    (\bm{I}-\bm{K}_k\bm{H}_k)
    \bm{P}^-_k
    (\bm{I}-\bm{K}_k\bm{H}_k)^\top
    +
    \bm{K}_k\bm{\Sigma}_m\bm{K}_k^\top,
    \label{eq:ekf_covariance_correction}\\
    \bm{H}_k
    &=
    \begin{bmatrix}
        \bm{I}_3 & \bm{0} & \bm{0} & \bm{0} & \bm{0}\\
        \bm{0} & \bm{0} & \bm{I}_3 & \bm{0} & \bm{0}
    \end{bmatrix}.
\end{align}
The quaternion correction is applied multiplicatively as
\begin{equation}
    \hat{\bm{q}}_k
    =
    \mathcal{N}_q\left(
        \operatorname{Exp}_q(\delta\hat{\bm{\theta}}_k)
        \otimes
        \hat{\bm{q}}^-_k
    \right),
    \label{eq:ekf_quaternion_correction}
\end{equation}
while the remaining components are corrected additively. The state supplied to the NMPC is finally constructed as
\begin{equation}
    \hat{\bm{x}}_k
    =
    \begin{bmatrix}
        (\hat{\bm{p}}^n_k)^\top &
        \hat{\bm{q}}_k^\top &
        \left(
            \bm{R}^{\top}(\hat{\bm{q}}_k)
            \hat{\bm{v}}^n_k
        \right)^\top &
        \left(
            \tilde{\bm{\omega}}^b_k-\hat{\bm{b}}_{g,k}
        \right)^\top
    \end{bmatrix}^{\top}.
    \label{eq:ekf_nmpc_state}
\end{equation}

\subsection{Reference Generation and Docking}

For both navigation and docking, RRT* \cite{karaman2011sampling} generates a geometric path $\mathcal{P}=\{\bm{p}^{r}_0,\ldots,\bm{p}^{r}_{M}\}$ from the current vehicle position to the desired terminal position. The path is interpolated and time-parametrized at constant speed to obtain $\mathcal{X}^r_k$ in \eqref{eq:reference_sequence}. During docking, the RRT* goal is the predefined dock position $\bm{p}^n_d$, while the attitude reference is interpolated from the initial attitude to $\bm{q}_d$. The perception module also supports ArUco-based relative dock-pose estimation \cite{aruco}.

\subsection{Nonlinear Model Predictive Control}

At prediction stage $\ell$, the quaternion tracking error is
\begin{align}
    \tilde{\bm{q}}_{\ell|k}
    &=
    \bm{q}^{r}_{k+\ell|k}
    \otimes
    \bm{q}^{-1}_{k+\ell|k},
    \nonumber\\
    \bm{e}_{q,\ell|k}
    &=
    \operatorname{sgn}
    \left(
        \tilde{q}_{w,\ell|k}
    \right)
    \operatorname{vec}
    \left(
        \tilde{\bm{q}}_{\ell|k}
    \right),
    \label{eq:quaternion_tracking_error}
\end{align}
where $\operatorname{vec}(\cdot)$ extracts the vector part of a quaternion. The sign correction selects the shortest quaternion representation and avoids unwinding. The stage cost is then defined as
\begin{align}
    &\ell(\bm{x}_{\ell|k},\bm{\tau}_{\ell|k})
    ={}
    \left\|
        \bm{p}^{n}_{\ell|k}
        -
        \bm{p}^{n,r}_{k+\ell|k}
    \right\|_{\bm{Q}_p}^{2}
    +
    \left\|
        \bm{e}_{q,\ell|k}
    \right\|_{\bm{Q}_q}^{2}
    \nonumber\\
    &+
    \left\|
        \bm{v}_{\ell|k}
        -
        \bm{v}^{r}_{k+\ell|k}
    \right\|_{\bm{Q}_v}^{2}
    +
    \left\|
        \bm{\omega}_{\ell|k}
        -
        \bm{\omega}^{r}_{k+\ell|k}
    \right\|_{\bm{Q}_{\omega}}^{2}
    +
    \left\|
        \bm{\tau}_{\ell|k}
    \right\|_{\bm{R}_{\tau}}^{2},
    \label{eq:nmpc_stage_cost}
\end{align}
where, for any vector $\bm{e}$ and weighting matrix $\bm{W}$, $\|\bm{e}\|_{\bm{W}}^2:=\bm{e}^{\top}\bm{W}\bm{e}$. The terminal cost $V_f(\bm{x}_{N|k})$ contains the same state-tracking terms as \eqref{eq:nmpc_stage_cost}, weighted by $\bm{P}_p$, $\bm{P}_q$, $\bm{P}_v$ and $\bm{P}_{\omega}$, respectively, without an input penalty.


Finally, the NMPC problem solved at time step $k$ is
\begin{equation}
\begin{aligned}
    \min_{\substack{
        \bm{x}_{0|k},\ldots,\bm{x}_{N|k}\\
        \bm{\tau}_{0|k},\ldots,\bm{\tau}_{N-1|k}
    }}
    \quad&
    \sum_{\ell=0}^{N-1}
    \ell(\bm{x}_{\ell|k},\bm{\tau}_{\ell|k})
    +
    V_f(\bm{x}_{N|k})
    \\
    \mathrm{s.t.}\quad&
    \bm{x}_{0|k}=\hat{\bm{x}}_k,
    \\
    &
    \bm{x}_{\ell+1|k}
    =
    \bm{f}_d(
        \bm{x}_{\ell|k},
        \bm{\tau}_{\ell|k}
    ),
    \\
    &
    \underline{\bm{\tau}}
    \leq
    \bm{\tau}_{\ell|k}
    \leq
    \overline{\bm{\tau}},
    \quad
    \ell=0,\ldots,N-1.
\end{aligned}
\label{eq:nmpc_ocp}
\end{equation}

With only nonempty input-box constraints, recursive feasibility follows directly \cite{rawlings2020model}. The nonlinear model and cost functions are implemented symbolically using CasADi \cite{andersson2019casadi}. The Optimal Control Problem (OCP) is solved using acados \cite{acados}, with a Gauss-Newton Hessian approximation, four-stage Explicit Runge-Kutta integration, partial-condensing HPIPM and a Sequential Quadratic Programming Real-Time Iteration (SQP-RTI) scheme. The solution from the previous time step is shifted to warm-start the optimization. At each sampling time, only the first optimized body wrench $\bm{\tau}_{0|k}^{\star}$ is sent to PX4 and the problem is solved again using the updated state estimate and reference. All the controller, reference generation and EKF code is released open-source\footnote{\url{https://github.com/KTH-DHSG/bluerov2_control}}.


\section{Experimental Evaluation}
\label{sec:experiments}

\subsection{Experimental Setup}

The physical experiments are conducted in the Marinarium using the modified BlueROV2 Heavy and the docking station shown in Fig.~\ref{fig:docking_and_simulation}. The underwater motion-capture system provides global pose measurements, while the vehicle IMU provides acceleration and angular-velocity measurements to the EKF. The water current $\bm{v}^{n}_{c}=\bm{0}$ is assumed negligible.

All estimation, planning and control algorithms are executed onboard. The NMPC operates with sampling time $T_s=\SI{0.04}{s}$ and horizon $N=25$, i.e., a prediction interval of $\SI{1}{s}$. The terminal state weights are set to twice the stage weights, with $|\bm{f}|\leq[44,44,13.7]^\top\,\si{N}$ and $|\bm{m}|\leq[30,16.5,21]^\top\,\si{N.m}$.

All the experimental parameters are summarized in Table~\ref{tab:experimental_parameters}. The perception, state-estimation and NMPC modules are evaluated using the available underwater data and the autonomous docking experiment. Their computational loads are reported separately, while the aggregate onboard utilization is estimated from the module-wise measurements.

\begin{table}[t]
    \centering
    \caption{Experimental and algorithmic parameters.}
    \label{tab:experimental_parameters}
    \scriptsize
    \begin{tabular}{c|l}
        \hline
        Parameter & Value\\
        \hline
        Camera profiles
        &
        RGB 1280x720x15fps; D 848x480x30fps
        \\
        YOLO input resolution
        &
        1280x720 pixels
        \\
        YOLO confidence threshold
        &
        0.9
        \\
        Jetson power mode
        &
        15W
        \\
        IMU sampling rate
        &
        200\,\si{Hz}
        \\
        Motion-capture rate
        &
        100\,\si{Hz}
        \\
        $T_s$, $N$
        &
        $\SI{0.04}{s}$, $25$
        \\
        $\bm{Q}_p$, $\bm{Q}_q$
        &
        $\operatorname{diag}(15,100,50)$,
        $\operatorname{diag}(5,5,5)$
        \\
        $\bm{Q}_v$, $\bm{Q}_{\omega}$
        &
        $\operatorname{diag}(5,70,30)$,
        $\operatorname{diag}(1,1,1)$
        \\
        $\bm{R}_{\tau}$
        &
        $\operatorname{diag}(0.1,0.1,0.1,0.05,0.05,0.05)$
        \\
        \hline
    \end{tabular}
\end{table}

\subsection{Underwater Perception and Onboard Performance}

\begin{figure*}[t]
    \centering
    \begin{subfigure}[t]{0.35\textwidth}
        \centering
        \includegraphics[
            width=\linewidth
        ]{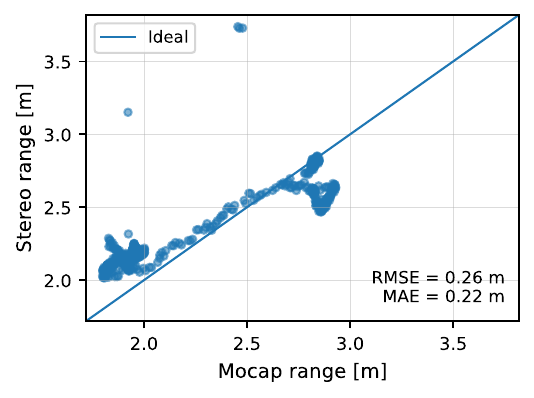}
        \caption{Stereo-ranging error.}
        \label{fig:perception_results_depth}
    \end{subfigure}
    \hfill
    \begin{subfigure}[t]{0.37\textwidth}
        \centering
        \includegraphics[
            width=\linewidth
        ]{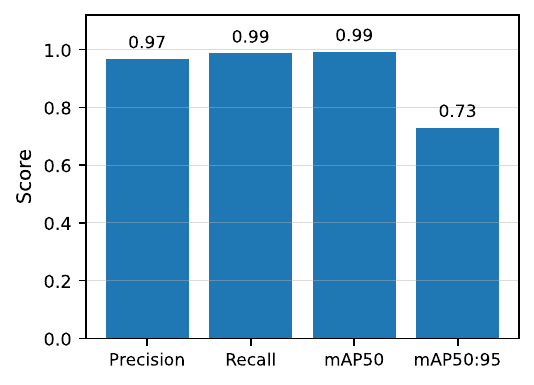}
        \caption{BlueROV2 detection performance.}
        \label{fig:perception_results_yolo}
    \end{subfigure}
    \hfill
    \begin{subfigure}[t]{0.254\textwidth}
        \centering
        \includegraphics[
            width=\linewidth
        ]{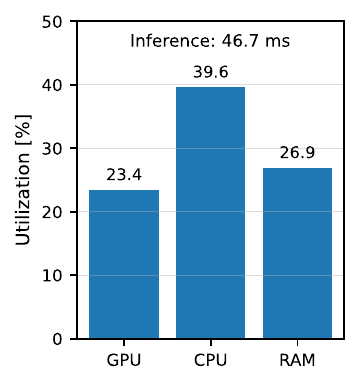}
        \caption{Computational performance.}
        \label{fig:perception_results_compute}
    \end{subfigure}
    \caption{Perception evaluation: (a) stereo range versus motion-capture ground truth; (b) held-out YOLO precision, recall and mAP; and (c) Jetson Orin NX utilization and inference time for the selected profile.}
    \label{fig:perception_results}
\end{figure*}

The stereo-ranging evaluation uses an underwater recording of a BlueROV2 observed at reference distances between 1.8 and 2.9\,\si{m}. The distances are obtained using a docked BlueROV2 as the target object and another BlueROV2 with the camera as the observer. The estimated range in \eqref{eq:detected_object_depth} is compared with the motion-capture ground truth. For $n_z$ accepted measurements, the depth Root Mean Square Error (RMSE) is
\begin{equation}
    e_{\mathrm{RMSE}}^{z}
    =
    \sqrt{
        \frac{1}{n_z}
        \sum_{k=1}^{n_z}
        \left(
            \hat{z}_{k}-z^{\mathrm{ref}}_{k}
        \right)^2
    }.
    \label{eq:depth_rmse}
\end{equation}
The Mean Absolute Error (MAE) is also reported in Fig.~\ref{fig:perception_results_depth}, together with the stereo and motion-capture ranges, compared with the ideal line of error-free measurements.

The object detector is evaluated on 1067 labeled underwater BlueROV2 images, using separate recorded sequences for training, validation, and testing. Precision, recall, $\mathrm{mAP}_{50}$ and $\mathrm{mAP}_{50:95}$ are reported in Fig.~\ref{fig:perception_results_yolo}. Figure~\ref{fig:perception_results_compute} reports the inference time and CPU, GPU and memory utilization during onboard execution on the Jetson Orin NX.

The final selected operating profile uses a color resolution of 1280x720 at 15\,\si{Hz} and a depth resolution of 848x480 at 30\,\si{Hz}. It achieves a depth RMSE of 0.26\,\si{m}, a valid-depth proportion of 55.38\,\% and a detector rate of 15\,\si{Hz}.

\begin{figure*}[t]
    \centering
    \begin{subfigure}[t]{0.31\textwidth}
        \centering
        \includegraphics[
            width=\linewidth
        ]{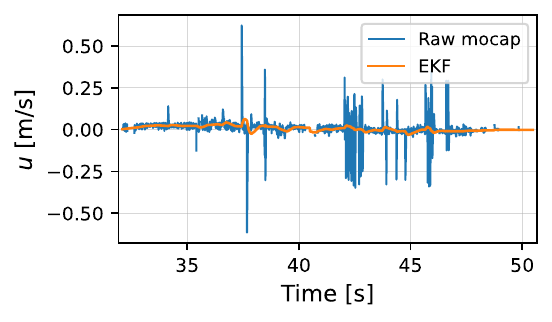}
        \caption{Raw and EKF-filtered surge velocity.}
        \label{fig:control_results_ekf}
    \end{subfigure}
    \hfill
    \begin{subfigure}[t]{0.31\textwidth}
        \centering
        \includegraphics[
            width=\linewidth
        ]{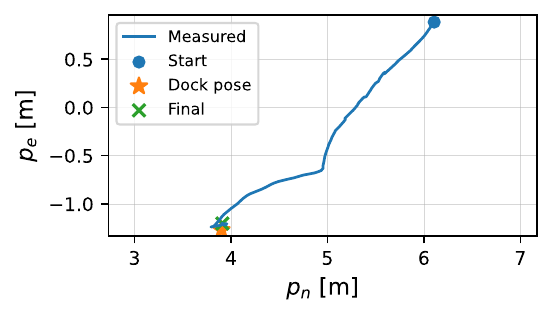}
        \caption{Measured trajectory and docking pose.}
        \label{fig:control_results_trajectory}
    \end{subfigure}
    \hfill
    \begin{subfigure}[t]{0.36\textwidth}
        \centering
        \includegraphics[
            width=\linewidth
        ]{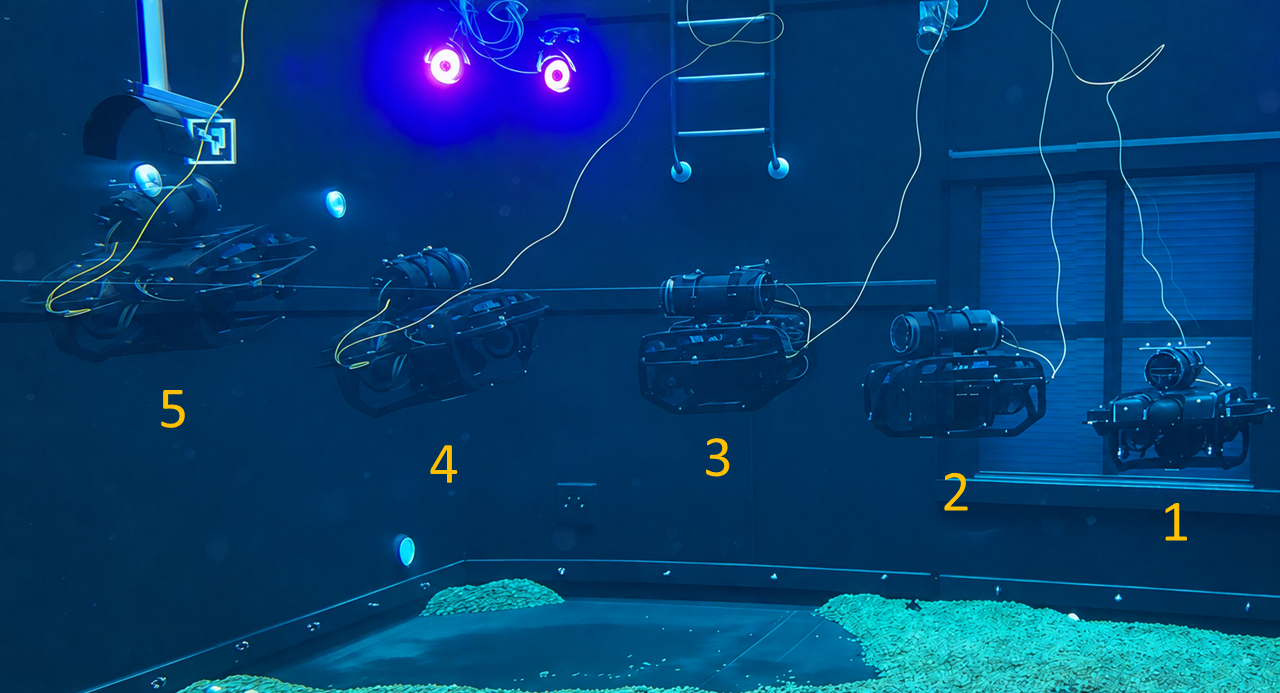}
        \caption{Docking sequence.}
        \label{fig:control_results_docking}
    \end{subfigure}
    \caption{Docking evaluation: (a) raw motion-capture and EKF surge velocity during the final hold; (b) measured horizontal trajectory, commanded dock pose and final position; and (c) overlaid frames of the physical docking sequence.}
    \label{fig:control_results}
\end{figure*}

\subsection{State Estimation and NMPC Tracking}

The EKF is evaluated using the autonomous docking experiment. Since the motion-capture stream contains several tracking losses, including a maximum interruption of approximately 1.19\,\si{s}, the analysis focuses on measurement filtering and robustness to losses rather than absolute localization accuracy. Figure~\ref{fig:control_results_ekf} compares the raw motion-capture odometry with the filtered estimate. During the final hold interval, the EKF reduces the body-velocity standard deviation by approximately 74\%.

The NMPC tracking performance is evaluated during the docking using the filtered state estimate. Figure~\ref{fig:control_results_trajectory} shows the measured trajectory with the commanded docking pose.

\subsection{Autonomous Docking}

The vehicle follows the time-parametrized RRT* reference to the predefined docking pose, with a speed of \SI{0.10}{m/s}. The trajectory lasts \SI{32.1}{s}, after which the NMPC maintains the final reference. The final position and attitude errors are \SI{0.067}{m} and \SI{10.9}{deg}, respectively. Figure~\ref{fig:control_results_docking} shows the final docking configuration (see accompanying \href{https://youtu.be/VsCCucKPt8k}{video}).


\section{Conclusion} \label{sec:conclusion}
In this work, we developed an open BlueROV2 Heavy autonomy platform combining underwater stereo perception, quaternion-based state estimation, reference planning and 6 DoF NMPC for navigation and docking. Open hardware, ROS~2/PX4 software, Gazebo environments, multi-robot simulation tools and a low-cost docking station support reproducible experimentation. Experiments demonstrate onboard relative perception, state estimation, NMPC tracking and autonomous docking. Future work investigates sampling-based methods, e.g., Model Predictive Path Integral control, safety-critical multi-robot coordination and diver-robot interaction.


\bibliographystyle{IEEEtran}
\bibliography{Reference}


\end{document}